\documentclass[12pt,twocolumn,letterpaper]{article}
\usepackage{cvpr}
\usepackage{graphicx}
\usepackage{hyperref}
\usepackage{tabularx}
\usepackage{times}
\usepackage{epsfig}
\usepackage{graphicx}
\usepackage{amsmath}
\usepackage{amssymb}
\usepackage[font=small,labelfont=bf]{caption}
\usepackage{cite}
\usepackage[flushleft]{threeparttable}

\newcommand*\rot{\rotatebox{90}}

\newcolumntype{s}{>{\hsize=1.25\hsize}X}

\usepackage{titlesec}
\titleformat{\section}{\normalfont\Large\bfseries}{\thesection}{1em}{}
\titleformat{\subsection}{\normalfont\large\bfseries}{\thesubsection}{1em}{}
\titleformat{\subsubsection}{\normalfont\normalsize\bfseries}{\thesubsubsection}{1em}{}

\cvprfinalcopy 

\def\cvprPaperID{****} 

\title{\vspace{-0.5cm}2DDATA: 2D Detection Annotations Transmittable Aggregation \\ for Semantic Segmentation on Point Cloud}

\author{Guan Cheng Lee\\
National Cheng Kung University, Taiwan\\
{\tt\small nm6111027@gs.ncku.edu.tw}
}

\begin{document}

\maketitle

\begin{abstract}
    Recently, multi-modality models have been introduced because of the complementary information from different sensors such as LiDAR and cameras. 
    It requires paired data along with precise calibrations for all modalities, the complicated calibration among modalities hugely increases the cost of collecting such high-quality datasets,
    and hinder it from being applied to practical scenarios. Inherit from the previous works, we not only fuse the information from multi-modality without above issues, and also exhaust the information in the RGB modality
    We introduced the 2D Detection Annotations Transmittable Aggregation(\textbf{2DDATA}),
    designing a data-specific branch, called \textbf{Local Object Branch}, which aims to deal with points in a certain bounding box, because of its easiness of acquiring 2D bounding box annotations.
    We demonstrate that our simple design can transmit bounding box prior information to the 3D encoder model,
    proving the feasibility of large multi-modality models fused with modality-specific data.
\end{abstract}


\begin{figure*}[!htb]
    \minipage{0.32\textwidth}
    \centering
      \includegraphics[width=\linewidth]{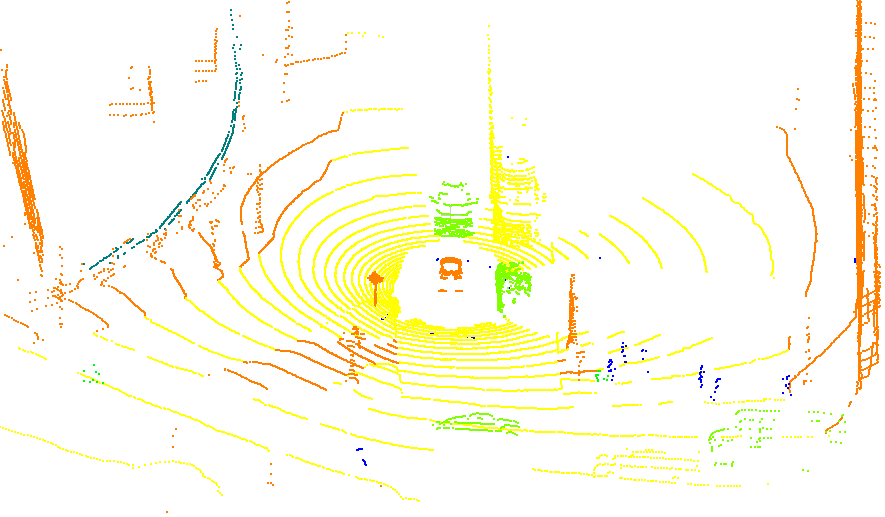}
      (a) Full Point Cloud Scene \label{fig:pc_cls}
    \endminipage\hfill
    \minipage{0.32\textwidth}
    \centering
      \includegraphics[width=\linewidth]{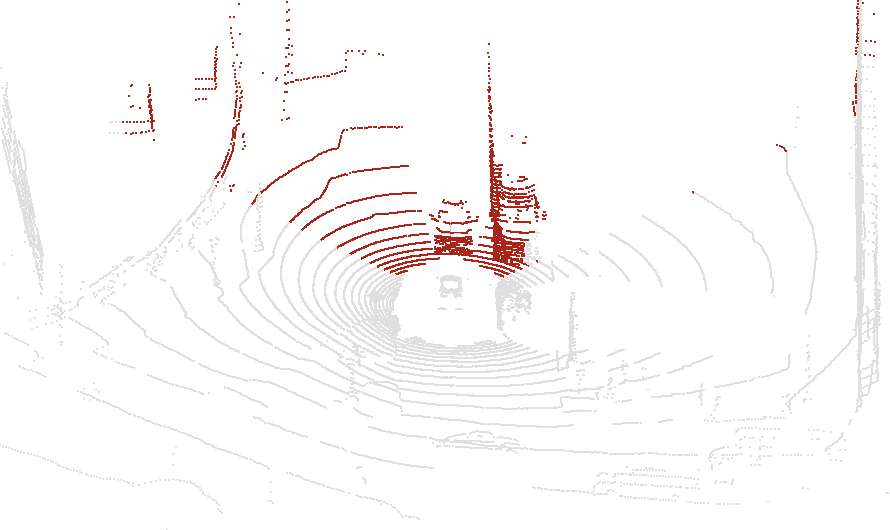}
      (b) Points in a Camera View \label{fig:pc_fov}
    \endminipage\hfill
    \minipage{0.32\textwidth}%
    \centering
      \includegraphics[width=\linewidth]{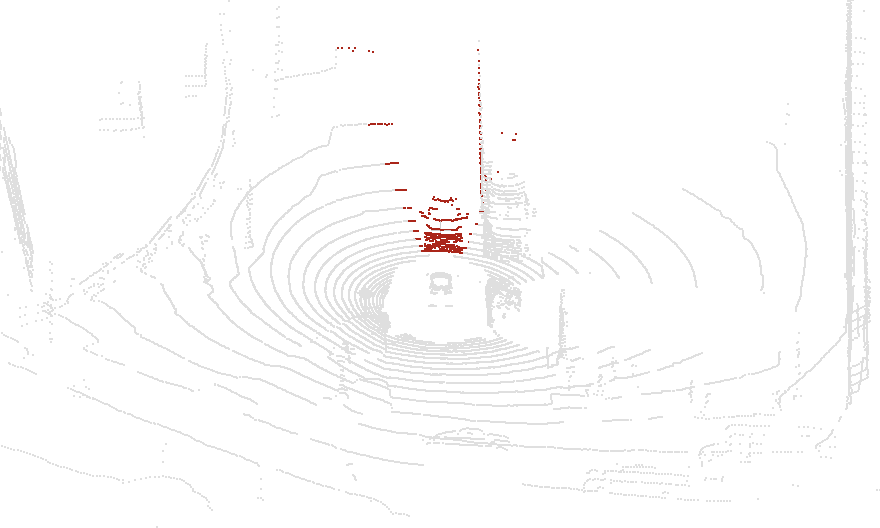}
      (c) Points in a Bounding Box \label{fig:pc_inbox}
    \endminipage
    \caption{Point Cloud projection view. This figure shows the different scale of levels of data,
            for 2DPASS, the 3D encoder model predicts all points in (a), and the 2D-3D fusion model predicts
            all the points in (b). 2DDATA dives further into the object level, magnify and transits the information in bounding boxes.
                }
\end{figure*}

\section{Introduction}
The rapid development of robust intelligent systems such as self-driving cars requires high-performance models to illustrate the 3D environments more precisely and accurately.
Based on Point Cloud, hence, the Lidar segmentation has become one of the most crucial topics recently. Also, the pure LiDAR solution is gradually not compatible to the complicated behavior of systems and increasing security concerns,
the multi-modality becomes one of most potential works. However, the more modalities are introduced,
the model would be inevitably becoming more massive and the calibrations among modalities are also more complicated, both inhibits it from being applied to practical scenarios.

Even though 2DPASS \cite{yan20222dpass} demonstrated the effectiveness of learning between modalities, and remains the 3D encoder model to be minimum during inference,
 this method still requires a large amount of high-quality data for training, particularly for LiDAR, 
and the collection of corresponding paired images exacerbates this issue.
Building on the success of \cite{yan20222dpass}, we explore the potential of knowledge distillation in the context of multi-modality. By collecting more information from one existing modality, in our case, 2D bounding boxes in the camera, we expect 
the prior knowledge of bounding boxes is transmitted to the 3D encoder model, just like the role of 2D branch in \cite{yan20222dpass}, since the 2D bounding box is quite easy to collect compared to LiDAR-image fused data. 

Based on this motivation, we propose a modality data-specified architecture, 2D Detection Annotations Transmittable Aggregation (\textbf{2DDATA}),
to prove that providing the information in one modality in multi-modality is also a feasible way to improve the main 3D encoder model.

Our contributions are concluded as follows:
\begin{itemize}
    \item We propose a simple adaptive architecture to prove that by providing 2D image annotations, the 3D encoder model can be improved without adding any inference cost.
    \item We provide novalty ideas about the multi-modality model, which is not only about the fusion of different modalities but also the fusion of different data levels, which is potentially useful for future research.
\end{itemize}

\section{Related Work}
\subsection{Pure LiDAR Solutions}
Because of the conventional understanding, people are used to aligning the training and the inference data of deep learning models,
makes the LiDAR-only input the most common way for Point Cloud Segmentation. Even though the limiting data compared to multi-modality,
these solutions \cite{tang2020searching,kong2023rethinking,zhu2021cylindrical,cheng20212, qiu2022gfnet, ye2022efficient} still achieves great results.
Most of the contributions focused on improving the effectiveness of 3D models, SPVCNN \cite{tang2020searching} is one of the important foundation works recently,
it equips Sparse Convolution to encode voxel-level and also point-level with MLP layers, becomes one of efficient SOTA of pure LiDAR models.
\cite{cheng20212} design a U-Net like architecture, the combinations of Sparse Convolution and the attention mechanism make it to be competent for SPVCNN.

Besides the above researches, some of them try to figure out other ways to re-encode the coordinates of point cloud data.
Cylinder3D \cite{zhu2021cylindrical} treats point cloud as the cylindrical partition compared to the cubic partition in voxel-level view.
GASN \cite{ye2022efficient} leveraged the computation cost and gained great improvement by considering the sparsity of the voxels. Another similar thought is to project points onto different views, like Bird's-eye View(BEV) \cite{zhang2020polarnet}, Range view \cite{kong2023rethinking} or their combinations \cite{qiu2022gfnet}.

However, the single-modality model inevitably suffered from the amount of information in the data,
the disadvantages are even deteriorating because of the dramatically increasing amount of applicable multi-sensors data recently.

\subsection{Multi-Sensor Solutions}
For the reason that the dramatical increasing of the amount of data and modalities,
multi-modality becomes another popular solutions to 3D applications.
Because of The complementary properties of modalities, it significantly boosts the performance of 3D segmentation models,
these solutions \cite{zhuang2021perception,liu2022bevfusion,liang2022bevfusion,li2022deepfusion,li2023logonet,yan20222dpass} are dedicated to close the gap between modalities with reasonable and efficient ways.
PMF \cite{zhuang2021perception} is one of the starting research, but it requires the paired image and point cloud
even during inference time, which makes it almost impossible to deploy on real world applications.
There are similar to all works like \cite{liu2022bevfusion,liang2022bevfusion,li2022deepfusion,li2023logonet}, some of them even requires multi-camera images.

2DPASS \cite{yan20222dpass} debuted as a gamer changer, it brings distinguished improvement under the premise that
the multi-modality module only activates during the training process, which indicates we could obtain better 3D segmentation models
by designing a complex training process, without adding any computation cost during inference time.

However, 2DPASS is still constrained by the limiting modality data. It is still costly to collect high quality and well-calibrated multi-modality data,
we proposed 2DDATA to alleviate this problem by fully utilizing the rich information in RGB images with its 2D bounding box.

\subsection{Frustums on Point Cloud}
Since the inherent design of the camera pinhole model and 2D data is much easier to obtain, it is natural to investigate the properties of frustums on Point Cloud, and mainly,
much more 3D detection works are accomplished than 3D Segmentation, it might result from the similarity of the geometry between 2D and 3D detection works, In this subsection, we would provide researches about 3D Detection, this is still beneficial to our work.

By providing prior knowledge of 2D bounding box, some of the researches achieve acceptable results under designing constraints,
like Frustum pointnets \cite{qi2018frustum}, it produces Point Cloud by RGB-D data, and generate 2D bounding box by CNN, to predict 3D box. 
\cite{li2019gs3d} predict not only 2D box but also its orientation, and manage to close the gap between 2D proposal and 3D ground truth,
in this work, only proposed 2D box information and RGB image is input, there is no 3D data involved.
FGR \cite{wei2021fgr} is much closed to our work, with Point Cloud and 2D bounding box labels to craft the key point and edge of 3D box.
These researches provide the initial muses of our work.

\section{Technical Approach}
In this section, as our method heavily relies on the projection between 3D point cloud and 2D images,
we will provide a detailed explanation in the first subsection on how we match the 3D point cloud information with pixels of 2D color images.
In section 3.2, we will elaborate on the current SOTA architecture, 2DPASS.
Lastly, in section 3.3, we propose the improved network architecture 2DDATA,
to \textbf{magnify} and \textbf{transmit} the details of interesting objects given from the prior knowledge from 2D bounding box, improving the performance of the main 3D encoder model.
A detailed description of the system architecture and our local object branch will also present in section 3.3.

\subsection{Preliminaries}
The collection of 3D point cloud data relies on LiDAR sensor, which is a device uses light to objects or surfaces,
by collecting, calibrating and calculating the reflected light, gathering the information including
the world coordinates $p_i = (x_i, y_i, z_i) \in \mathbb{R}^3 $ and reflection rate, for each $i$ represents a point in Point Cloud.
Camera is normally used during the data collection process, it would simultaneously take RGB pictures along with LiDAR working,
we could define the image data as $pix_i=(u_i, v_i)$, for each pixel in the image,
to be more specific, $[u_i, v_i]$ represents the x and y of each pixel on the image, like $[u_0, v_0] = [0, 0]$, $[u_0, v_1] = [0, 1]$, $[u_1, v_0] = [1, 0]$ etc.

Besides the above two kinds of data, we also have the transformation parameters,
these parameters enable us to rotate and shift from one coordinate system to another, for example,
by using the camera extrinsic matrix, we could transform a 3D point from the LiDAR coordinate to the camera coordinate.

\begin{figure}
    \includegraphics[width=\linewidth]{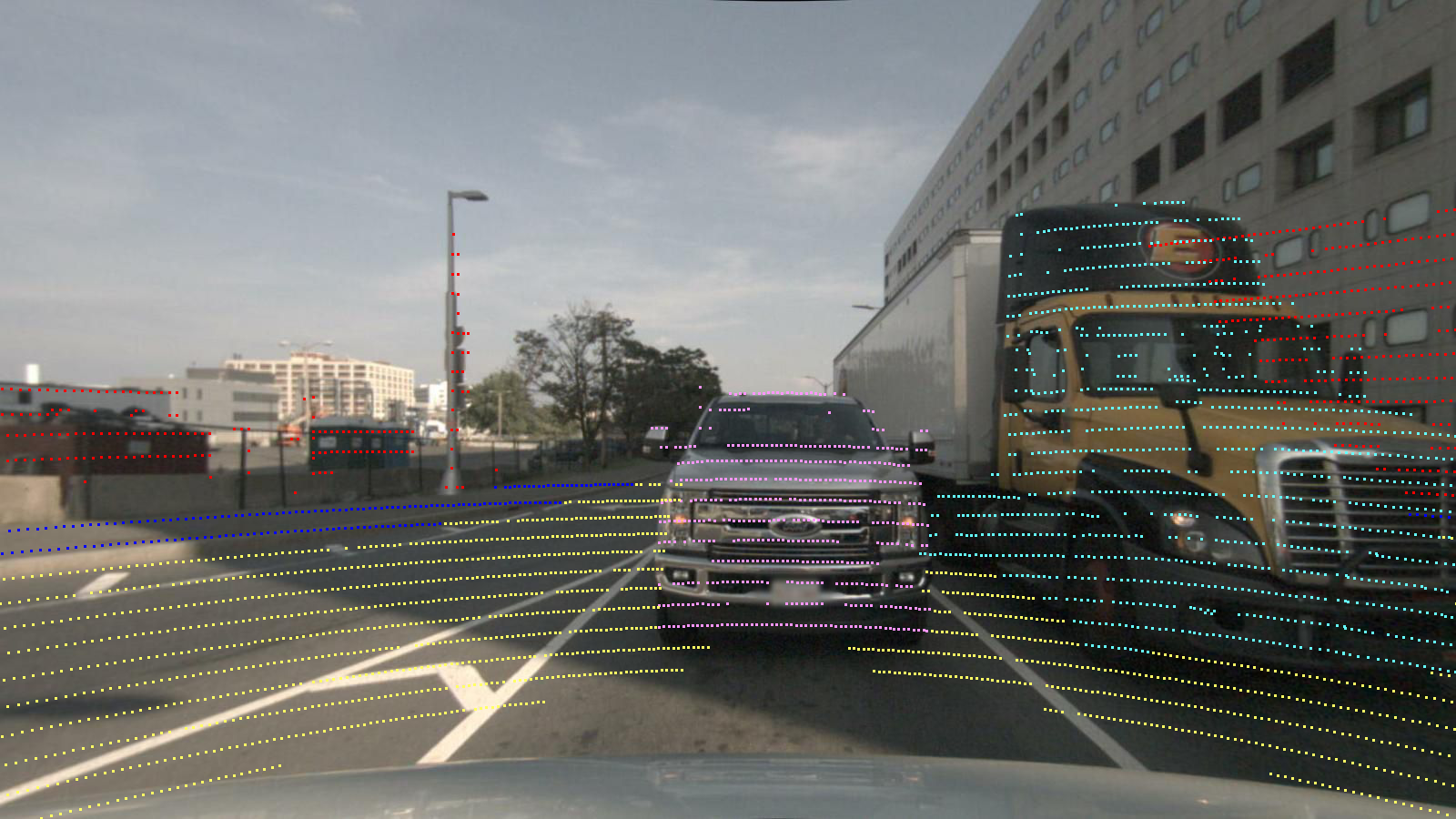}
    \captionsetup{belowskip=-10pt}
    \caption{NuScenes projection example.
    The different color of these points represent the category they belong to,
    there are car, truck, drivable surface, sidewalk and manmade in this image, respectively.}
    \label{fig:nuScenes_projection_example}
\end{figure}

\subsubsection*{Coordinate Transformation and Rasterization}
For a given LiDAR point cloud $P=\{p_i\}_{i=1}^{N}$ and a given image plane $I=\{{{pix}_i}\}_{i=1}^{M}$, where $N$ is the number of points in the point cloud scene, M is the number of pixels of the image.
Assuming Point Cloud and RGB images are captured at the exactly same time, the objective is to rasterize the 3D points $p_i$ onto ${pix}_i$ 
by using a series of matrices multiplying, the general camera calibration formula is given as:
\begin{equation}
    [u_i, v_i, 1]^T = \frac{1}{z_i} \times K \times T \times [x_i, y_i, z_i, 1]^T ,
\end{equation}
where $K \in \mathbb{R}^{3 \times 4}$ is the camera intrinsic matrix 
and $T \in \mathbb{R}^{4 \times 4}$ is the camera extrinsic matrix,
the $T$ transforms the world coordinates into camera coordinates, the K transforms the camera coordinates to pixel coordinates.

In SemanticKITTI\cite{behley2019semantickitti} dataset, $K$ and $T$ has been directly provided.
Because of the difference between the frequency of capturing data of LiDAR and camera in NuScenes \cite{caesar2020nuscenes}, we need to consider the timestamp of the extrinsic matrices, the $T$ is given as:
\begin{flalign}
\begin{aligned}
    T=T_{camera \gets ego_{t_c}} \times T_{ego_{t_c} \gets global} \\ \times T_{global \gets ego_{t_l}} \times T_{ego_{t_l} \gets lidar} ,
\end{aligned}
\end{flalign}
where the subscripts marks what the coordinates transform from and to, for instance,
$T_{camera \gets ego_{t_c}}$ is the transformation matrix to project the view of the vehicle position at timestamp $t_c$ to its camera view, and so on.

To generally speak, they serve the purpose of projecting 3D data between coordinate systems, that is camera, LiDAR, vehicle and world coordinates.
The above transformation helps us to convert the LiDAR captured at timestamp $t_l$ into the view of camera at timestamp $t_c$.
The projection example is showed at \hyperref[fig:nuScenes_projection_example]{Fig. 2}

\begin{figure}
    \includegraphics[width=\linewidth]{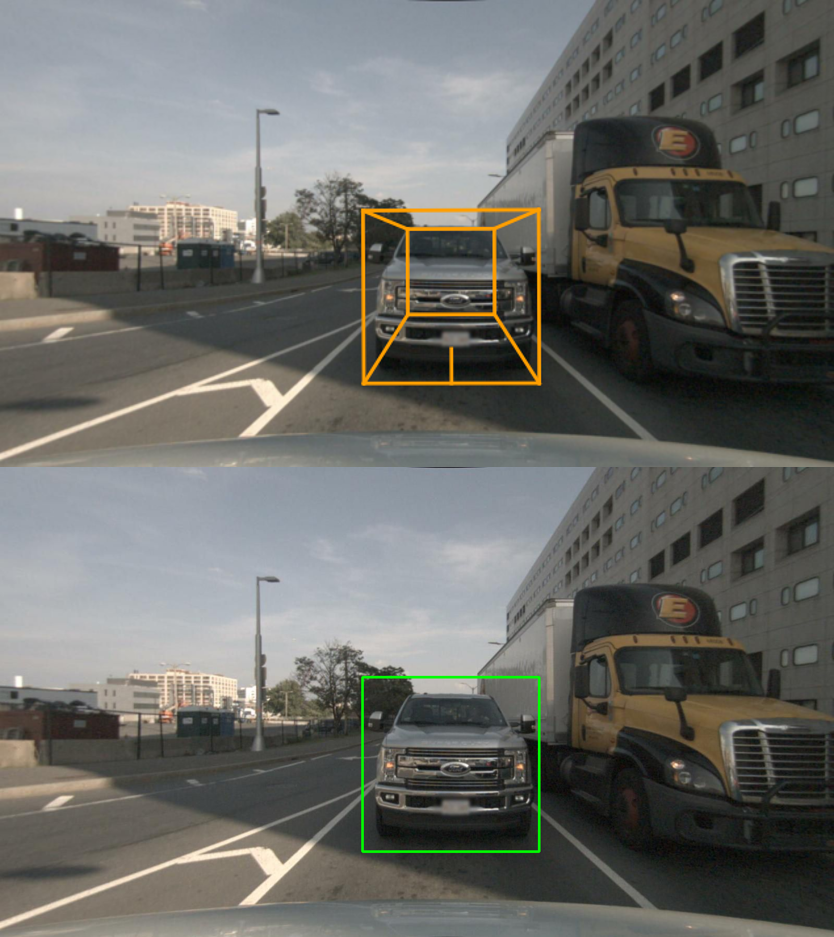}
    \captionsetup{belowskip=-10pt}
    \caption{The top image is the 3D bounding box from NuScenes dataset,
             by projecting the cube, we would have 8 points on the image plane,
             then choose 4 points that make the 2D bounding box cover the largest
             area to form the corresponding 2D bounding box.}
    \label{fig:nuScenes_box_acquirement_projection_example}
\end{figure}

\subsubsection*{Point-to-Pixel Matching}
After projection, we would have a set of points $p_{i}^{\prime} \in I$, they are projected $p_i$ in the camera FOV(field of views).
Notice that the points and pixels are not in a one-to-one relationship,
the number of points remains the same after projection, a pixel might contain more than one point, and there are some pixels has no points in it,
the point-to-pixel mapping could be written as:
\begin{equation}
    p_{i}^{\prime} = \{\lfloor u_{i}^{\prime} \rfloor, \lfloor v_{i}^{\prime} \rfloor \}, \; for \; i=1,...,N^{img}
\end{equation}

where $\lfloor \cdot \rfloor$ is the floor operation since the projected $u_{i}^{\prime},  v_{i}^{\prime}$ are float, and $N^{img}$ is the number of points projected onto the image plane, normally, $N > N^{img}$. With the above mapping, we could find the corresponding image features for a given point $p_i$.

\subsubsection*{Data Acquirements}
Because NuScenes \cite{caesar2020nuscenes} dataset does not provide 2D but 3D bounding box annotations, we follow the above process to project
3D bounding box onto the image plane, therefore for each 3D box, we would have 8 points on the image plane, then the largest width and height
is chosen to acquire a 2D bounding box. The projection example is showed at \hyperref[fig:nuScenes_box_acquirement_projection_example]{Fig. 3}

\subsection{2DPASS Overview}
\begin{figure}
    \includegraphics[width=\linewidth]{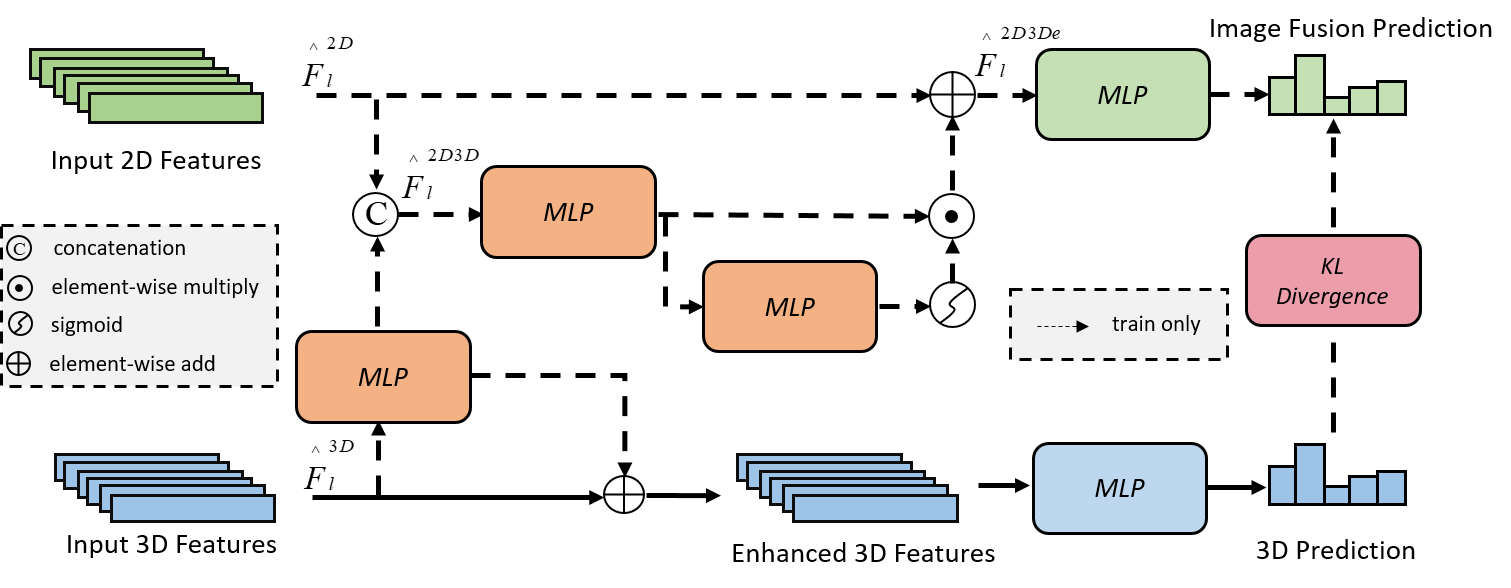}
    \captionsetup{belowskip=-10pt}
    \caption{Part of the architecture of 2DPASS. This figure shows how the 2DPASS transfer the image modal and the disposable training schema.
    During inference, only 3D branch (downside of the figure) is conducted the segmentation.}
    \label{fig:2DPASSarch}
\end{figure}

The success of 2DPASS \cite{yan20222dpass} inspired us to investigate other possibilities of knowledge distillation training
without adding any cost during inference. There are three preferable properties is mentioned in its original paper:
\textbf{1) Generality:} Its 3D encoder model can be easily replaced with any other 3D segmentation models, by leveraging the latency and accuracy.
\textbf{2) Flexibility:} The fusion module is only used during training, makes the training design more flexible.
\textbf{3) Effectively:} By only adding a small section of overlapped multi-modality data, the method boosts the performance of the 3D segmentation model remarkably.
Considering the great importance of 2DPASS for our work, we will go through the details of 2DPASS in this section.
Its architecture is showed at \hyperref[fig:2DPASSarch]{Fig. 4}.

\subsubsection*{Modality Fusion}
For a given pair of a 2D image and 3D point cloud, 
2DPASS extracts 2D features $\hat{F}_{l}^{2D} \in \mathbb{R}^{N^{img} \times D_l}$ and 3D features $F_{l}^{3D} \in \mathbb{R}^{N \times D_l}$
by 2D encoder FCN\cite{long2015fully} and 3D encoder SPVCNN\cite{tang2020searching}, since they are both pyramid networks,
the $l$ represents the feature extracted from l-th layer, the $D_l$ is the number of channels of features at l-th layer.

Because we would like to focus on the points in the image FOV,
we first select the points in the image FOV to have $\hat{F}_{l}^{3D}  \in \mathbb{R}^{N^{img} \times D_l}$ in
the 3D features $F_{l}^{3D}$ which is from the 360\textdegree{} full scene,
each $\hat{F}_{l}^{3D}$ could find a corresponding 2D feature by the Point-to-Pixel Matching.

$\hat{F}_{l}^{3D}$ are then input into a MLP layer, called \textbf{2D Learner}, the purpose of 2D Learner
is to narrow the gap between 2D and 3D features, after this, they are concatenated as fused features, written as:

\begin{equation}
    \hat{F}_{l}^{2D3D} = concat(\hat{F}_{l}^{2D}, 2DLearner(\hat{F}_{l}^{3D}))
\end{equation}

then knowledge distillation (KD) design \textbf{Multi-Scale Fusion-to-Single Knowledge Distillation} is introduced.

\subsubsection*{Multi-Scale Fusion-to-Single Knowledge Distillation (MSFSKD)}
\textbf{MSFSKD} consists of a few MLP layers, and formed as an attention-like module, which
weights its own features are beneficial to the predictions

\begin{equation}
    \hat{F}_{l}^{2D3D_e} = \hat{F}_{l}^{2D} + \sigma{(MLP(\hat{F}_{l}^{2D3D}))} \odot MLP(\hat{F}_{l}^{2D3D}),
\end{equation}

where $\sigma$ denotes Sigmoid activation function and $\odot$ is the element-wise multiplying.
After a simple MLP as a classifier, $\hat{F}_{l}^{2D3D_e}$ is input into final Fusion Prediction $S_{l}^{2D3D} \in \mathbb{R}^{N^{img} \times c}$, c is the number of classes.

\subsubsection*{Loss}
2DPASS simply apply the general loss for semantic segmentation tasks, which consists of cross-entropy loss and Lovasz losses\cite{berman2018lovasz}.

Another important contribution is introducing knowledge distillation via KL divergence, the distillation loss $L_{xM}$ as follows:
\begin{equation}
    L_{xM} = D_{KL}(S_{l}^{2D3D} || S_{l}^{3D}),
\end{equation}

the $S_{l}^{3D}$ is the segmentation predictions of the image scene from $\hat{F_l^{3D}}$, the above loss enforces the uni-directional distillation by pushing $S_{l}^{3D}$ closer to $S_{l}^{2D3D}$ ,
which results in the knowledge transfer.

\subsection{2DDATA Overview}
In 2DPASS \cite{yan20222dpass}, the RGB image features enable the 3D encoder model to understand the geography better via
visual information. However, the RGB image can be crafted for inquiring more 2D information. Therefore, we introduce \textbf{2DDATA} to take it further by providing 2D bounding boxes.

To involve the prior knowledge of bounding boxes, we design the \textbf{Local Object Branch}.
Put it simply, the 2DDATA architecture is built upon original 2DPASS architecture, replacing MSFSKD with Local Object Branch.
The architecture of 2DDATA is showed at \hyperref[fig:2DDATAarch]{Fig. 5}.

\subsection{Local Object Branch}
\begin{figure}
    \includegraphics[width=\linewidth]{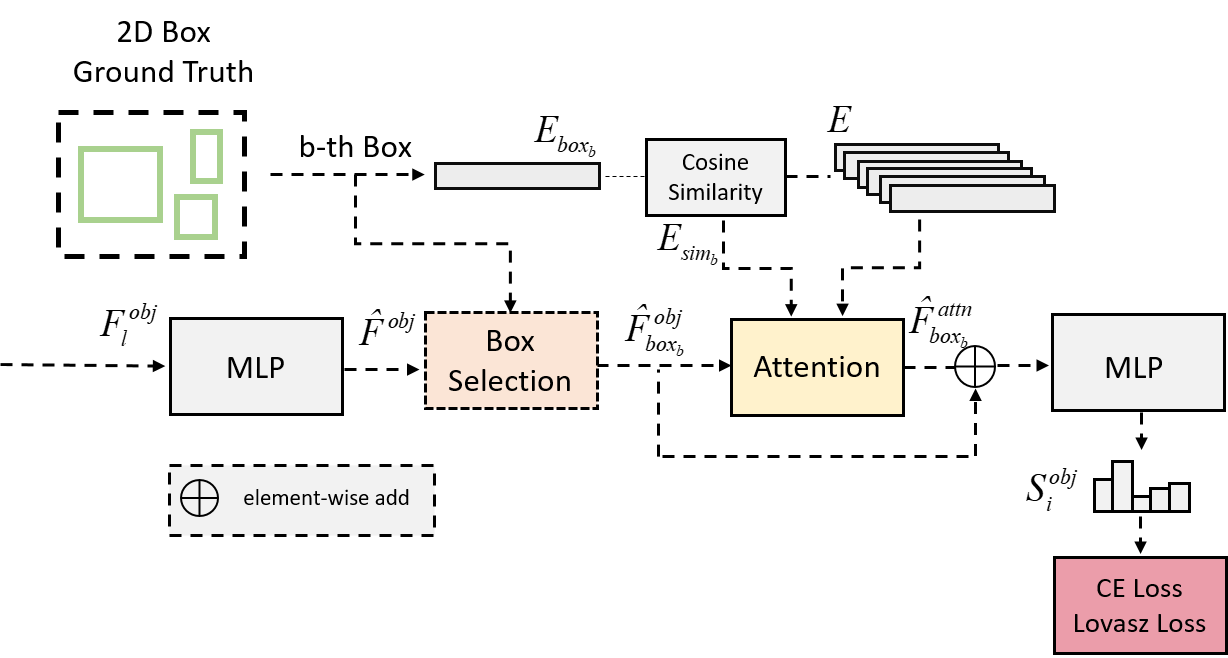}
    \captionsetup{belowskip=-10pt}
    \caption{Local Object Branch.}
    \label{fig:LOB}
\end{figure}

Local Object Branch is designed to deal with the points in a specific 2D bounding box,
the 2D bounding box is a strong prior that indicate the points in the region is highly possible belonging to 
the class of bounding box labeled. With this extra information, the module should first consider to classify the points 
as the same class to bounding box. By leveraging the geographic relationship and image features, this should be remarkably easy
to recognize the points in the bounding box. (\hyperref[fig:LOB]{Fig. 5}) shows the architecture of Local Object Branch.

\subsubsection*{Fused and Box-selected Features}
The 3D features in the image scope $\hat{F}_{l}^{3D}$ would pass to a MLP layers, just like the 2D Learner in \cite{yan20222dpass},
the learned 3D features would be concatenated with 2D features, so we would have:

\begin{equation}
    \hat{F}_{l}^{obj} = concate(MLP_l(\hat{F}_{l}^{3D}), \hat{F}_{l}^{2D}),
\end{equation}
note that the 2D learner is a specific MLP for each l-th features.We further concatenate features along the l-th layer, and downsample it back to hidden size as the modal-fusion features $\hat{F}^{obj} \in \mathbb{R}^{N^{img} \times D}$, which can be written as:
\begin{equation}
    \hat{F}^{obj} = MLP(concate(\{\hat{F}_{l}^{obj}\}_{i=1}^{L})),
\end{equation}
We could find the features of points $\hat{F}_{box_b}^{obj} \in \mathbb{R}^{N^{box_b} \times D}$ located in the bounding box $b$, note that the different bounding box contains different number of points $N^{box_b}$:
\begin{multline}
    \hat{F}_{box_b}^{obj} = \\ \{\hat{f}_i|f_i \in \hat{F}^{obj}, x_1 < p_{i, 1}^{\prime} < x_2, y_1 < p_{i, 2}^{\prime} < y_2 \}_{i=1}^{N^{box_b}},
\end{multline}
where $x1, x2, y1, y2$ represents the boundary of the bounding box $b$, the $N^{box_b}$ is the number of points in the bounding box, and $N > N^{img} > N^{box_b}$.

\begin{figure}[t]
    \includegraphics[width=\linewidth]{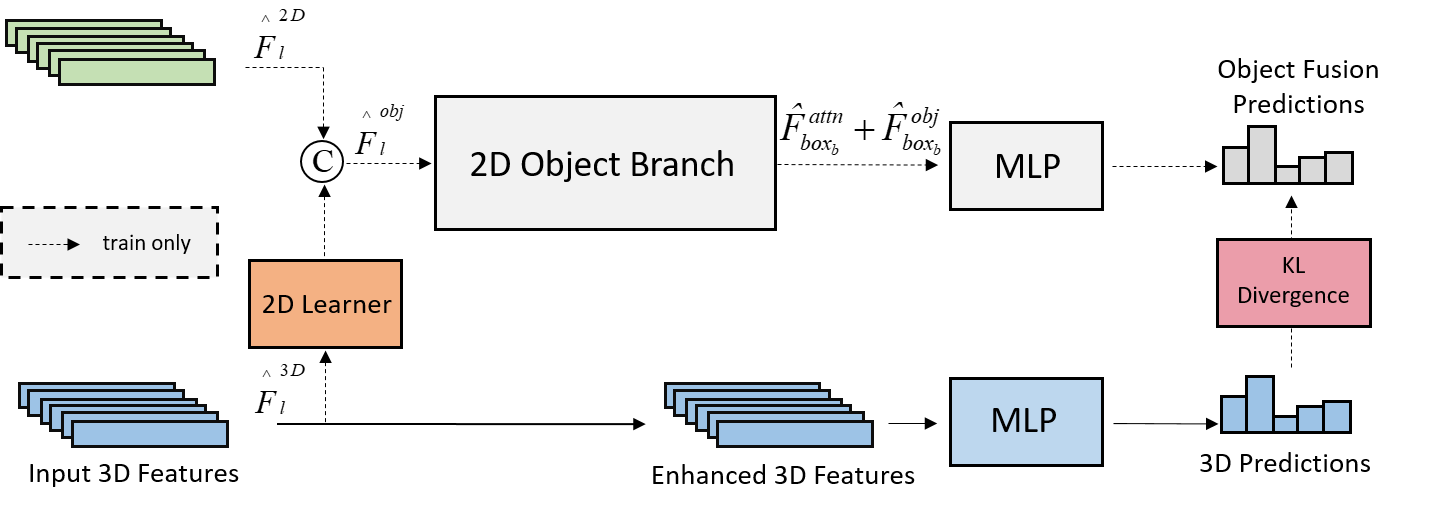}
    \captionsetup{belowskip=-10pt}
    \caption{2DDATA architecture.
    We build 2DDATA upon original 2DPASS, replace MSFSKD with Local Object Branch.}
    \label{fig:2DDATA}
\end{figure}

\subsubsection*{Class Aware Attention}
To involve the prior knowledge of bounding box, the \textbf{Class Aware Attention} is designed.
We create a bag of class embeddings vectors $E_{class} = \{E_i\}_{i=1}^{num\_class} \in \mathbb{R}^{c \times D}$, when there are $N_{1}$ points belonging to $class1$,
we would have $N_1$ number of $E_1$, for $N^{box_b}$ points and a given bounding box $b$, we obtain their class relationship between each point and the box $b$ via consine similarity:

\begin{equation}
    E_{sim_b} = \dfrac{ E \cdot E_{box_b}}{max(\lVert E \rVert \cdot \lVert E_{box_b} \rVert, \epsilon)},
\end{equation}

where $E \in \mathbb{R}^{N^{box_b} \times D}$ is the sets of class embeddings of points, and $E_{box_b}$ represents the class embeddings of each point and the bounding box $b$, respectively.

$E_{sim_b} \in \mathbb{R}^{N^{box_b}}$ means the cosine similarity of the class of the bounding box $b$ and the class of points in $b$.
This similarity would be the key vectors of attention block,
and $\hat{F}_{box_b}^{obj}$ as the value vectors, $E$ as the query vectors, the Class Aware Attention can be written as:
\begin{equation}
    \hat{F}_{box_b}^{attn} = concat(head_1, \dots ,head_h),
\end{equation}
and the i-th head:
\begin{equation}
    head_i = softmax(\frac{EW^Q (E_{sim_b}W^K)^T}{\sqrt{D}})\hat{F}_{box_b}^{obj}W^V,
\end{equation}

where $W^K \in \mathbb{R}^{1 \times D} $ and $W^Q, W^V \in \mathbb{R}^{D \times D}$ are the learnable parameters, $D$ is the hidden size of the attention block.

With Class Aware Attention, we could encode the class information of the bounding box into Local Object Branch,
this simple design proves the Scalability of the stacking design of 2DDATA.

\begin{table*}[ht]
    \label{tab:benchmark}
    \centering
    \small
    \begin{threeparttable}
    \begin{tabularx}{\textwidth}{l|c|sssssssssssssssss}
    \hline
      Method & \rot{\textbf{mIoU}} & \rot{barrier} & \rot{bicycle} & \rot{bus} & \rot{car} 
      & \rot{construction} & \rot{motorcycle} & \rot{pedestrian} & \rot{traffic cone} 
      & \rot{trailer} & \rot{truck} & \rot{driveable} & \rot{sidewalk} 
      & \rot{terrain} & \rot{manmade} & \rot{vegetation} \\
      \hline
      \hline
      2DPASS       & 72.87 & \bf{69.2} & 37.4 & \bf{93.7} & \bf{88.8} & 47.0 & \bf{82.9} & 77.5 & 57.9 & 50.0 & \bf{81.4} & \bf{95.7} & 71.5 & \bf{71.5} & 87.0 & \bf{85.3} \\
      \hline
      2DDATA& 73.28 & 69.0 & 36.9 & 92.3 & 87.8 & 49.7 & 82.1 & \bf{77.6} & \bf{58.7} & 52.9 & 79.7 & \bf{95.7} & \bf{71.9} & 71.4 & 87.0 & 85.2 \\
      \hline
      2DDATA& \bf{73.33} & 69.0 & \bf{39.0} & 91.7 & 87.9 & \bf{50.4} & 81.5 & \bf{77.6} & 58.2 & \bf{55.0} & 81.1 & 95.6 & 71.5 & 71.4 & 87.0 & 84.4 \\
      \hline
    \end{tabularx}
    \caption{Semantic segmentation results on the NuScenes test set.}
    \end{threeparttable}
\end{table*}

\subsubsection*{Loss}
The loss of 2DDATA is pretty much the same as \cite{yan20222dpass}, we use the same segmentation loss for the predictions of Local Object Branch,
denoted as $S_{b}^{obj}$. 

\begin{equation}
    L_{seg} = CE(S_{b}^{obj} || GT_{b}^{obj})+Lovasz(S_{b}^{obj} || GT_{b}^{obj}),
\end{equation}

And we inherit the knowledge distillation mechanism of 2DPASS, transmit the information of Local Object Branch with KL divergence:
\begin{equation}
    L_{xM}^{obj} = D_{KL}(S_{b}^{obj} || S_{l}^{3D}),
\end{equation}
as above, we actually compare the predictions in each box to each layer of 3D predictions,
this result in the aggregation for all scales and each positions of features.

The Total loss of 2DDATA is:
\begin{equation}
    L_{total} = L_{seg} + \lambda L_{xM}^{obj},
\end{equation}
where $\lambda$ is the weight of the knowledge distillation loss, we set it to be 0.1 in 2DDATA.

\section{Experiments}
\subsection{Experiment Setups}
\subsubsection*{NuScenes Datasets}
Following the previous 3D segmentation works, we conduct experiments on NuScenes dataset.

NuScenes dataset is a large-scale and prominent dataset with various tasks for autonomous driving,
it contains 1000 scenes which consists of images, point cloud, calibration and annotation information from 1 LiDAR,
5 Radar and 6 camera etc., eligible to prove the generalizability of our proposed method. 

The 1000 scenes are divided into 700 training scenes, 150 validation scenes and 150 test scenes,
because of the long training process, we choose 277 out of 700 scenes as our training set against 150 validation set.
These 277 scenes is chosen based on the quality of the 3D bounding boxes,
at least one 3D bounding box of interest class is labeled as clearly visible in the scene.
We pick adult, child, bicycle, car, truck, motorcycle as our interest class in NuScenes.

\subsubsection*{Training Settings}
Both of the trainings are conducted 80 epochs, 2DDATA with batch size 12, learning rate 0.15, cosine warm-up scheduler 10 epoch is applied,
for 2DPASS, batch size 8 and learning rate 0.24 is applied, our method and 2DPASS both trained on the 277 scenes.

\subsection{Benchmark Results}
The results are reported at \hyperref[tab:benchmark]{Tab.\ 1}. The mIoU and overall accuracy are calculated over all 17 classes in NuScenes,
our method slightly outperforms 0.46 mIoU.

We would like to point out that Local Object Branch is sized 4.4 million parameters, while MSFSKD is reported with size 2.2 million parameters,
compared to the 3D encoder SPVCNN sized 45.6 million and 2D encoder FCN sized 26.4 million parameters, the Local Object Branch is relatively small, and just slightly larger than MSFSKD,
indicated that we are not introducing much extra parameters to get better results. Nevertheless, because of the inherited 2DPASS architecture, training cost is not our main concern.

\subsubsection*{Class Specific IoU}
We also present the class IoU, interestingly, the result is counter-intuitive. We observe that most of the mIoU of the interested class lower than 2DPASS method,
but others are hugely improved, like the construction is 3.4 higher and the trailer is 5.0 higher, both are not included in picked class bounding boxes.

We believe the reason is the competition between Local Object Branch and 3D encoder, the former "replaced" the ability of the latter to some extent,
which cause the performance of the interested class on the 3D encoder lower than 2DPASS method, increasing the weights of KL divergence loss might be an option to solve it. The further analysis is required to prove this hypothesis, we leave it as future work.

\section{Future Work}
In this paper, we show the other possibilities to improve 3D multi-modality segmentation models, since
we could easily have high quality 2D annotations data nowadays, there are at least three
ways to refine this work:

\textbf{1) 2D Segmentation:} In this work, we use 2D bounding box because of the similarity, it can only provide one single label for all points in the area.
With 2D segmentation, the more fine-grained prior inforamtion are provided, the model is expected to be more powerful.

\textbf{2) Automated Annotations:} It is still costly to label 2D annotations when the dataset is huge. With pretraining models,
we could instantly produce 2D annotations but they might be biased, the impacts need to be further addressed.

\textbf{3) Other Data in a Specific Modality:} We focus on RGB image information above, but it is possible to use data existing in other model.
For example, 3D bounding box in LiDAR Point Cloud or 2D box of Bird's-eye View in Radar Point Cloud.

We believe that our work is especially suitable for applying in engineering scenarios,
lessen the problem of lack of data in real world applications.

{\small
\bibliographystyle{ieeetr}
\bibliography{2ddata}

\begin{thebibliography}{10}

\bibitem{zhuang2021perception}
Z.~Zhuang, R.~Li, K.~Jia, Q.~Wang, Y.~Li, and M.~Tan, ``Perception-aware
  multi-sensor fusion for 3d lidar semantic segmentation,'' in {\em Proceedings
  of the IEEE/CVF International Conference on Computer Vision},
  pp.~16280--16290, 2021.

\bibitem{tang2020searching}
H.~Tang, Z.~Liu, S.~Zhao, Y.~Lin, J.~Lin, H.~Wang, and S.~Han, ``Searching
  efficient 3d architectures with sparse point-voxel convolution,'' in {\em
  Computer Vision--ECCV 2020: 16th European Conference, Glasgow, UK, August
  23--28, 2020, Proceedings, Part XXVIII}, pp.~685--702, Springer, 2020.

\bibitem{kong2023rethinking}
L.~Kong, Y.~Liu, R.~Chen, Y.~Ma, X.~Zhu, Y.~Li, Y.~Hou, Y.~Qiao, and Z.~Liu,
  ``Rethinking range view representation for lidar segmentation,'' {\em arXiv
  preprint arXiv:2303.05367}, 2023.

\bibitem{zhu2021cylindrical}
X.~Zhu, H.~Zhou, T.~Wang, F.~Hong, Y.~Ma, W.~Li, H.~Li, and D.~Lin,
  ``Cylindrical and asymmetrical 3d convolution networks for lidar
  segmentation,'' in {\em Proceedings of the IEEE/CVF conference on computer
  vision and pattern recognition}, pp.~9939--9948, 2021.

\bibitem{cheng20212}
R.~Cheng, R.~Razani, E.~Taghavi, E.~Li, and B.~Liu, ``2-s3net: Attentive
  feature fusion with adaptive feature selection for sparse semantic
  segmentation network,'' in {\em Proceedings of the IEEE/CVF conference on
  computer vision and pattern recognition}, pp.~12547--12556, 2021.

\bibitem{qiu2022gfnet}
H.~Qiu, B.~Yu, and D.~Tao, ``Gfnet: Geometric flow network for 3d point cloud
  semantic segmentation,'' {\em arXiv preprint arXiv:2207.02605}, 2022.

\bibitem{ye2022efficient}
M.~Ye, R.~Wan, S.~Xu, T.~Cao, and Q.~Chen, ``Efficient point cloud segmentation
  with geometry-aware sparse networks,'' in {\em Computer Vision--ECCV 2022:
  17th European Conference, Tel Aviv, Israel, October 23--27, 2022,
  Proceedings, Part XXXIX}, pp.~196--212, Springer, 2022.

\bibitem{zhang2020polarnet}
Y.~Zhang, Z.~Zhou, P.~David, X.~Yue, Z.~Xi, B.~Gong, and H.~Foroosh,
  ``Polarnet: An improved grid representation for online lidar point clouds
  semantic segmentation,'' in {\em Proceedings of the IEEE/CVF Conference on
  Computer Vision and Pattern Recognition}, pp.~9601--9610, 2020.

\bibitem{liu2022bevfusion}
Z.~Liu, H.~Tang, A.~Amini, X.~Yang, H.~Mao, D.~Rus, and S.~Han, ``Bevfusion:
  Multi-task multi-sensor fusion with unified bird's-eye view representation,''
  {\em arXiv preprint arXiv:2205.13542}, 2022.

\bibitem{liang2022bevfusion}
T.~Liang, H.~Xie, K.~Yu, Z.~Xia, Z.~Lin, Y.~Wang, T.~Tang, B.~Wang, and
  Z.~Tang, ``Bevfusion: A simple and robust lidar-camera fusion framework,''
  {\em arXiv preprint arXiv:2205.13790}, 2022.

\bibitem{li2022deepfusion}
Y.~Li, A.~W. Yu, T.~Meng, B.~Caine, J.~Ngiam, D.~Peng, J.~Shen, Y.~Lu, D.~Zhou,
  Q.~V. Le, {\em et~al.}, ``Deepfusion: Lidar-camera deep fusion for
  multi-modal 3d object detection,'' in {\em Proceedings of the IEEE/CVF
  Conference on Computer Vision and Pattern Recognition}, pp.~17182--17191,
  2022.

\bibitem{li2023logonet}
X.~Li, T.~Ma, Y.~Hou, B.~Shi, Y.~Yang, Y.~Liu, X.~Wu, Q.~Chen, Y.~Li, Y.~Qiao,
  {\em et~al.}, ``Logonet: Towards accurate 3d object detection with
  local-to-global cross-modal fusion,'' {\em arXiv preprint arXiv:2303.03595},
  2023.

\bibitem{yan20222dpass}
X.~Yan, J.~Gao, C.~Zheng, C.~Zheng, R.~Zhang, S.~Cui, and Z.~Li, ``2dpass: 2d
  priors assisted semantic segmentation on lidar point clouds,'' in {\em
  Computer Vision--ECCV 2022: 17th European Conference, Tel Aviv, Israel,
  October 23--27, 2022, Proceedings, Part XXVIII}, pp.~677--695, Springer,
  2022.

\bibitem{qi2018frustum}
C.~R. Qi, W.~Liu, C.~Wu, H.~Su, and L.~J. Guibas, ``Frustum pointnets for 3d
  object detection from rgb-d data,'' in {\em Proceedings of the IEEE
  conference on computer vision and pattern recognition}, pp.~918--927, 2018.

\bibitem{li2019gs3d}
B.~Li, W.~Ouyang, L.~Sheng, X.~Zeng, and X.~Wang, ``Gs3d: An efficient 3d
  object detection framework for autonomous driving,'' in {\em Proceedings of
  the IEEE/CVF Conference on Computer Vision and Pattern Recognition},
  pp.~1019--1028, 2019.

\bibitem{wei2021fgr}
Y.~Wei, S.~Su, J.~Lu, and J.~Zhou, ``Fgr: Frustum-aware geometric reasoning for
  weakly supervised 3d vehicle detection,'' in {\em 2021 IEEE International
  Conference on Robotics and Automation (ICRA)}, pp.~4348--4354, IEEE, 2021.

\bibitem{behley2019semantickitti}
J.~Behley, M.~Garbade, A.~Milioto, J.~Quenzel, S.~Behnke, C.~Stachniss, and
  J.~Gall, ``Semantickitti: A dataset for semantic scene understanding of lidar
  sequences,'' in {\em Proceedings of the IEEE/CVF international conference on
  computer vision}, pp.~9297--9307, 2019.

\bibitem{caesar2020nuscenes}
H.~Caesar, V.~Bankiti, A.~H. Lang, S.~Vora, V.~E. Liong, Q.~Xu, A.~Krishnan,
  Y.~Pan, G.~Baldan, and O.~Beijbom, ``nuscenes: A multimodal dataset for
  autonomous driving,'' in {\em Proceedings of the IEEE/CVF conference on
  computer vision and pattern recognition}, pp.~11621--11631, 2020.

\bibitem{long2015fully}
J.~Long, E.~Shelhamer, and T.~Darrell, ``Fully convolutional networks for
  semantic segmentation,'' in {\em Proceedings of the IEEE conference on
  computer vision and pattern recognition}, pp.~3431--3440, 2015.

\bibitem{berman2018lovasz}
M.~Berman, A.~R. Triki, and M.~B. Blaschko, ``The lov{\'a}sz-softmax loss: A
  tractable surrogate for the optimization of the intersection-over-union
  measure in neural networks,'' in {\em Proceedings of the IEEE conference on
  computer vision and pattern recognition}, pp.~4413--4421, 2018.

\end{thebibliography}
}

\end{document}